\documentclass[conference]{IEEEtran}
\usepackage{amsmath,amssymb,amsfonts}
\usepackage{algorithmic}
\usepackage{graphicx}
\usepackage{colortbl}
\usepackage{tabularx}
\usepackage{graphicx}

\begin{document}

\begin{onecolumn}
© 2023 IEEE. Personal use of this material is permitted. Permission from IEEE must be obtained for all other uses, in any current or future media, including reprinting/republishing this material for advertising or promotional purposes, creating new collective works, for resale or redistribution to servers or lists, or reuse of any copyrighted component of this work in other works.
\end{onecolumn}
\newpage
\begin{twocolumn}

\title{Human-interpretable and deep features \\ for image privacy classification
}

\author{\IEEEauthorblockN{Darya Baranouskaya}
\IEEEauthorblockA{\textit{Centre for Intelligent Sensing} \\
\textit{Queen Mary University of London}\\
London, UK \\
d.baranouskaya@se22.qmul.ac.uk}
\and
\IEEEauthorblockN{Andrea Cavallaro}
\IEEEauthorblockA{\textit{Centre for Intelligent Sensing} \\
\textit{Queen Mary University of London}\\
London, UK \\
a.cavallaro@qmul.ac.uk}
}
\maketitle
\begin{abstract}
Privacy is a complex, subjective and contextual concept that is difficult to define. Therefore, the annotation of images to train privacy classifiers is a challenging task. In this paper, we analyse privacy classification datasets and the properties of controversial images that are annotated with contrasting privacy labels by different assessors. We discuss suitable features for image privacy classification and propose eight privacy-specific and human-interpretable features. These features increase the performance of deep learning models and, on their own, improve the image representation for privacy classification compared with much higher dimensional deep features.
\end{abstract}

\section{Introduction}
With the abundance of images shared online, the protection of the privacy of an individual is an increasingly important concern. It is therefore important to develop accurate and effective methods for privacy classification to support individuals in their choices about protecting privacy. However, privacy image classification involves subjective, abstract and context-dependent decision-making, which is challenging to automate. 

Various deep-learning solutions have been developed for image privacy classification~\cite{b1, b2, b3, b4, b5}. These solutions include convolution neural networks for feature extraction~\cite{b1}, graph-based modeling~\cite{b2, b3}, using automatically generated tags~\cite{b4}, or image captioning methods that identify important topics~\cite{b5}. However, most of these approaches are not interpretable. 

In this paper, we analyse human privacy labels, hypothesise why certain decision determinations were made and propose a set of interpretable features that account for these decisions. We analyse three selected properties of an image, which are the level of sensitivity of the content, the probability of the presence and number of people, and the information about the place. 
We also analyse the labels assigned by different assessors on an image privacy dataset and how they correlate with each other. We perform an analysis of image privacy inspired by the human decision-making process and propose eight human-interpretable features. Based on these features, we show that simple algorithms outperform algorithms built on high-dimensional features extracted by neural networks and improve existing pre-trained deep neural networks for privacy classification.

\section{Features for privacy classification}
We consider PrivacyAlert~\cite{b6} and PicAlert~\cite{b7} as datasets for binary privacy classification. PrivacyAlert contains 6,288 currently available images. PicAlert has 37,535 images posted on Flickr from January to April 2010. In addition to binary privacy labels, PicAlert provides the results of a user study for 5,030 images in which 63 assessors classified subsets of images in 5 classes: \textit{clearly private, private, undecidable, public, clearly public}, with 65\% of the images annotated by at least two assessors. We analyse both the PicAlert user-study with two or more assessors per image and the PrivacyAlert datasets, but training and evaluation of models are performed only on the PrivacyAlert dataset due to uncertainty in target labels for the PicAlert user-study. 

We analyse the users' disagreements on annotated images from the PicAlert user-study dataset as a marker of how the labels of one user differ from average labels. We define an image as {\em controversial} if it was labelled private by at least one user and public by at least one another user and if no privacy label (i.e.~private or public) was significantly prioritised (selected by more than 65\% of assessors). Of these controversial images, 45\% contain one person, 45\% capture two or more people, and 10\% have no people. In PicAlert, 63\% of private images contain people and almost all images containing some nudity are private (99.50\% of images with adult content are private).

We consider the presence of people, place information and sensitivity of the images important attributes in human privacy decisions. 
 Information about the {\em people} in the image is represented by the number of people and the probability of people being present in the image. To incorporate {\em scene} information, we account for feature denoting the probability of an image being outdoors and of a specific place.  
 
 Most private outdoor images with no people contain either personal textual information (car registration plate and documents, signs or symbols defining political, religious or other beliefs), violent or medical content (photo of outgrowth skin or injuries), parts of human bodies, tattoos, unrecognised people or personal belongings (house, house yard). The hypothesis that outdoor images with no people are public is true in 78.5\% of cases (Figure~\ref{Diagram}). While the number of people in an image is not that important for privacy, the existence of people in the image is. Images with one person are also of particular interest. Figure~\ref{Distribution of private images conditioned on n of people} shows the cumulative distribution of private images conditioned on the number of people in the image. 

\begin{figure}[t!]
    \centering
    \includegraphics[width=0.45\textwidth]{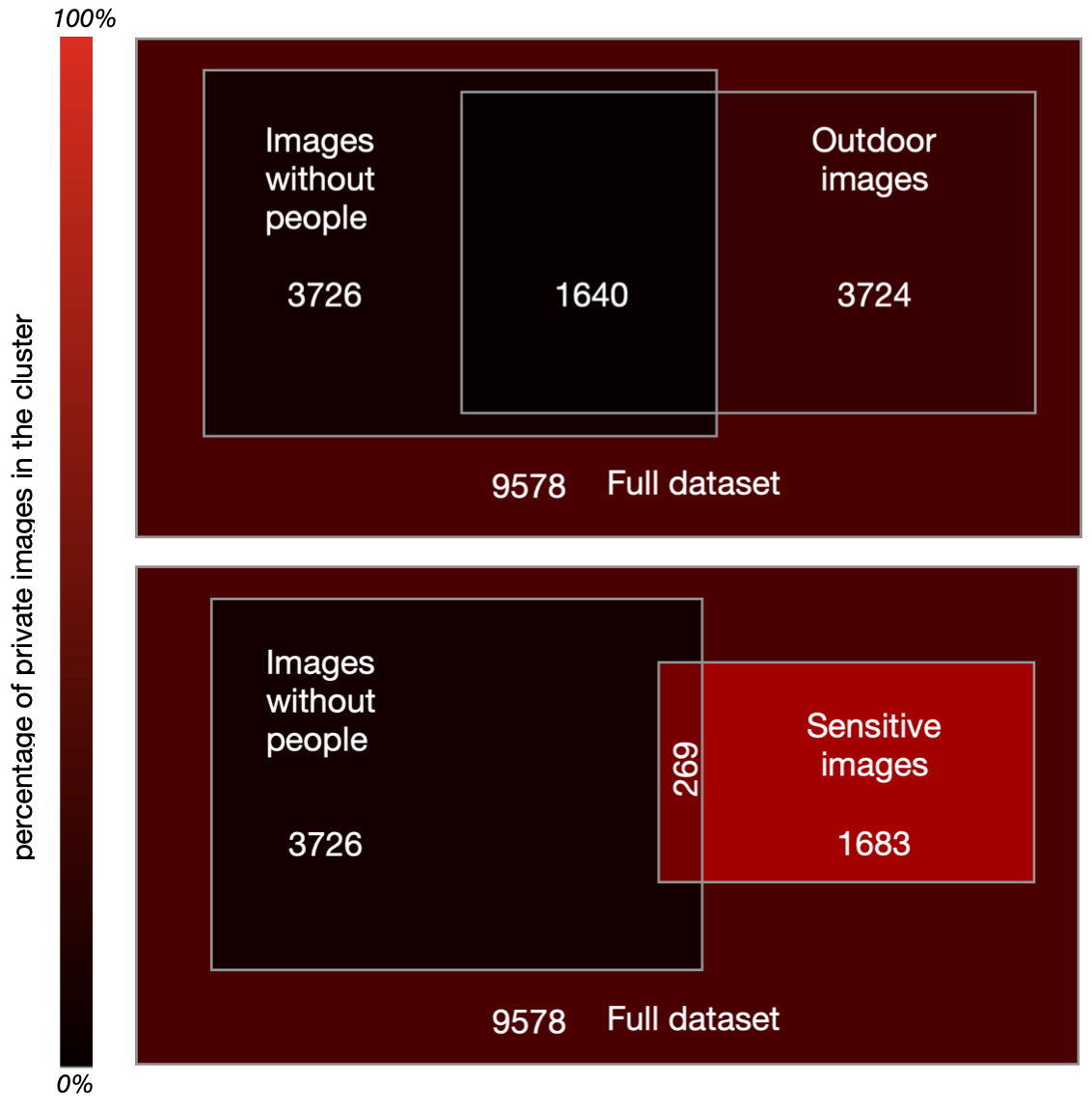}
    \caption{Distribution of images in the union of PicAlert user-study and PrivacyAlert datasets among groups of images including {\em images without people}, {\em sensitive images}, and {\em outdoor images} and probability of images belonging to groups being private.} 
    \vspace*{-4mm}
    \label{Diagram}
\end{figure}

Another concept we propose to analyse for privacy determination is the {\em sensitivity} of the image content. Our hypothesis is that sensitive images are more likely to be annotated as private and that the probability of an image belonging to one of the sensitive classes would be useful for privacy prediction. To this end, we consider the likelihood (5 scales from very unlikely to very likely) that the image contains \textit{adult} content (i.e.~explicitly sexual in nature, often with nudity and sexual acts);  \textit{racy} content  (i.e.~sexually suggestive but with less sexually explicit content than images tagged as adult, including strategically covered nudity, lewd or provocative poses, or close-ups of sensitive body areas); \textit{medical} content (i.e.~the interior of a body for clinical analysis and medical intervention); \textit{spoofed} content (i.e.~image modified to appear funny or offensive),  or \textit{violent} content.

Figure~\ref{Distribution of labels for sensitive features} shows the distribution of labels for each sensitive class for private and public images separately.  Only 17.57\% of all images are marked as sensitive (containing at least one type of sensitive content), where racy labels appear more often than others, in 16.13\% of the dataset, and only 3.15\% of all images are marked as adult, medical or violent. For every label of every sensitive class, we computed the probability of the image being private conditioned on the probability of the image belonging to the corresponding sensitive class (Figure~\ref{Probability of image being private conditioned on the probability of being sensitive}) and analysed image sensitivity together with the existence of people in the image.

Sensitive images are not necessarily private, although sensitive features, denoting the probability of a sensitive class for an image, are important for privacy classification. In the next section, we will show that the use of sensitive features improves the performance of simple privacy classifying models significantly.
\begin{figure}[t]
    \centerline{\includegraphics[width=0.5\textwidth]{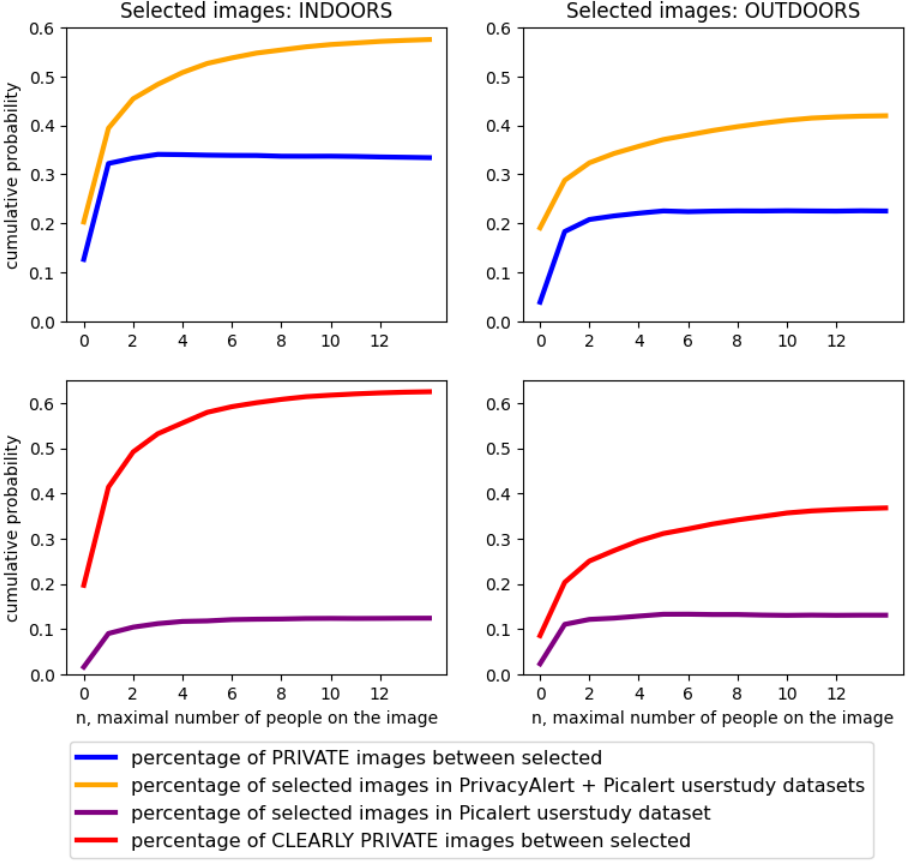}}
    \caption{Cumulative probability of an image being private as a function of the maximal number of people in an image.
    Left column: indoor images. Right column: outdoor images. }
    \vspace*{-4mm}
    \label{Distribution of private images conditioned on n of people}
\end{figure}

\begin{figure*}[t]
    \centering
    \includegraphics[width=0.94\linewidth]{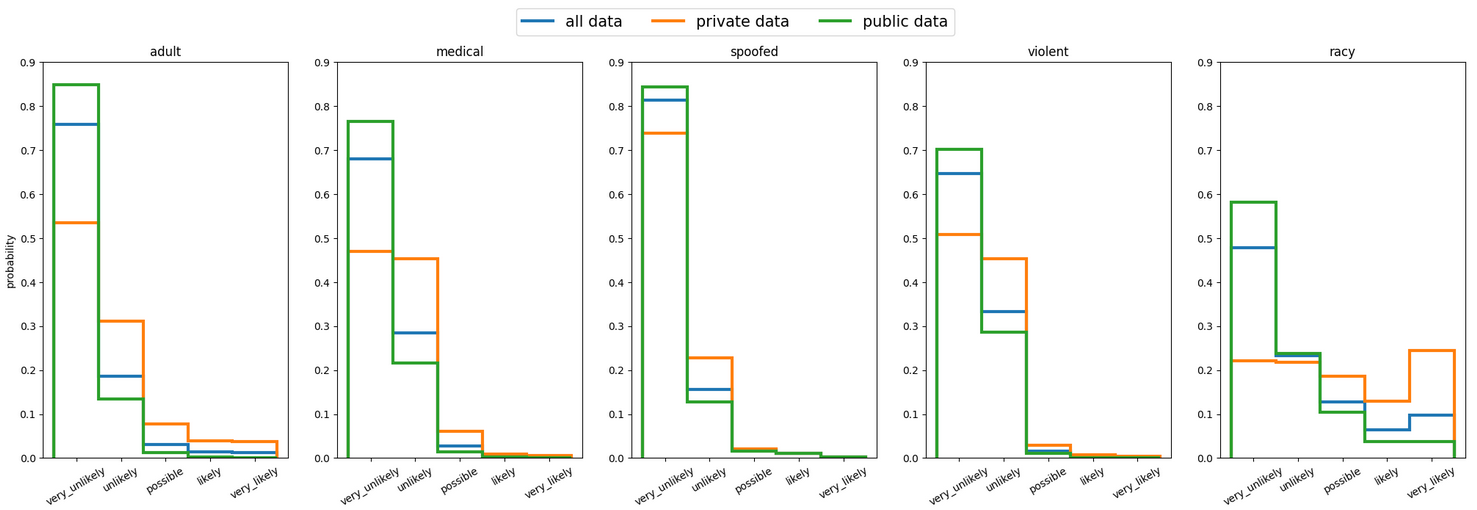}
    \caption{Distribution of labels for classes of sensitive content in images.}
    \vspace*{-4mm}
    \label{Distribution of labels for sensitive features}
\end{figure*}

\begin{figure}[t]
    \centering
    \includegraphics[width=0.48\textwidth]{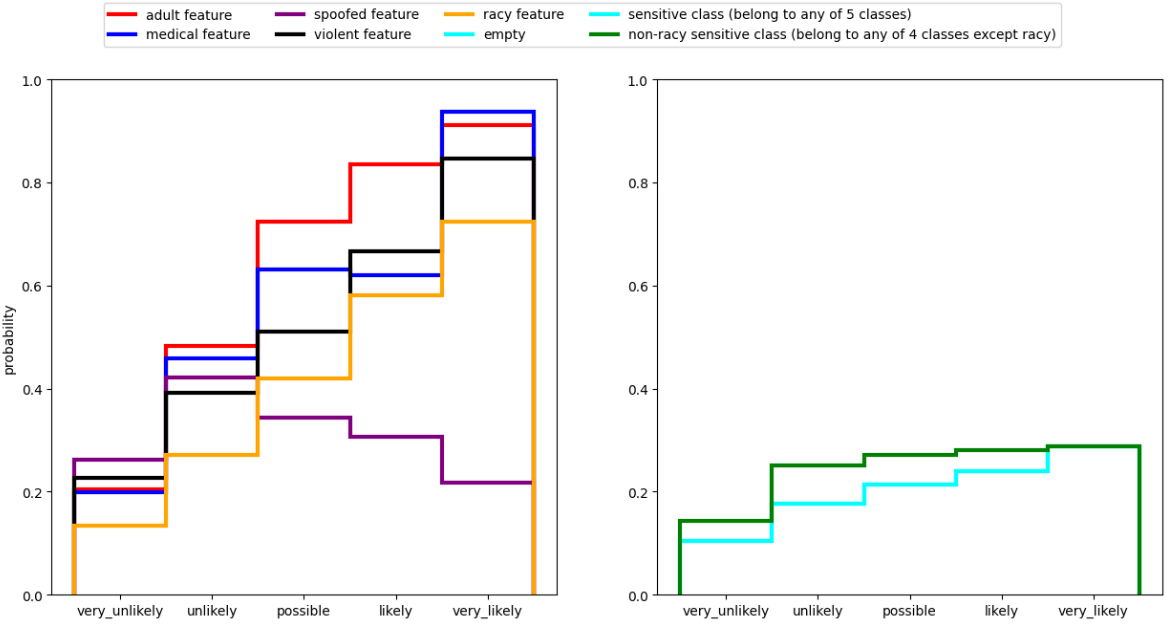}
    \caption{Probability of an image being private conditioned on the probability of belonging to the sensitive classes.} 
    \vspace*{-4mm}
    \label{Probability of image being private conditioned on the probability of being sensitive}
\end{figure}

\section{Analysis}
\label{validation}

In this section, we analyse deep features and the usefulness of human-interpretable features for privacy classification. We used Support Vector Machines (SVMs) operating on deep visual features derived from various layers of Convolutional Neural Networks (CNNs) because they have shown the best performance on privacy classification~\cite{b1}. In addition, we tested decision trees, logistic regressions (LogRegs) and multilayer perceptrons (MLPs) applied to features extracted from a pre-trained ResNet18, ResNet50 and ResNet101. As the combination of ResNet101 and MLP provided the best results, we use MLP as a reference. We also consider features extracted from Swin~\cite{b9} and ConvNeXt models~\cite{b10} for privacy classification, as vision transformers and modified CNNs have shown to outperform ResNets on classical computer vision tasks.  

For the specific implementation of this paper, to extract details about the {\em scene} we use a ResNet50  pre-trained on 365 scenes~\cite{b8} and  to extract the number and probability of people we use Yolo-v5~\cite{b11}. We extract sensitive features with the Google Cloud Vision SafeSearch API~\cite{b12}, which provides five levels of likelihood (\textit{very unlikely}, \textit{unlikely}, \textit{possible}, \textit{likely}, \textit{very likely}) for each of adult, racy, violent, spoofed and medical classes.

To evaluate the importance of specific features for privacy prediction, we trained several LogRegs on combinations of the following feature groups:  sensitive features (\textit{adult, medical, spoofed, violent, racy}), the existence of people (\textit{number of people, probability of people being in the image}), outdoors features (\textit{probability of an image being taken outdoors}), places 365 features (e.g.~\textit{airfield, alley, beauty salon}), 2,048 features from the second-last layer of ResNet101, 2,048 features from the second-last layer of ResNet50, and 512 features from the pre-last layer of ResNet18. 

Table~\ref{Performance of models trained on different subsets of features} shows that the  performance of LogReg on the eight privacy-specific features is only slightly worse than the performance of a LogReg trained on features extracted by ResNet101. Also using only eight privacy-specific features provides comparable performance to ResNet50 and better performance than using high-dimensional features extracted by Places365 and ResNet18 models. As baseline we combined LogReg with ResNet101 features, an approach inspired by~\cite{b1}, which showed the best deep-feature performance.

\begin{table}[t]
\caption{Performance of LogRegs trained on different subsets of features. Columns specify which features are included in the input feature vector for LogReg classification. Note that the ResNet features are not interpretable. Key -- BA: Balanced accuracy, F1: F1 score, Sens: sensitive features, People: existence of people, Out: outdoors, Places: features extracted by Places365,  ResNet: features extracted by ResNet18 (RN18), ResNet50 (RN50) or ResNet101 (RN101)}

\centering
\begin{tabular}{ c c c c c c c} 
\textbf{Sens} & \textbf{People} & \textbf{Out} & \textbf{Places} & \textbf{ResNet} &\textbf{\textit{BA}}& \textbf{\textit{F1}} \\
    \hline
    \checkmark &  &  &  &  & 80.04 & 66.46  \\
     & & & \checkmark & & 73.01 & 56.41  \\
     & & & & RN18 & 78.99 & 63.97  \\
     & & & & RN50 & 81.37 & 66.99 \\
     & & & & RN101 & 81.51 & 67.46  \\
    
    \hline
     \checkmark & \checkmark &  &  &  & 81.23 & 67.76 \\
     & \checkmark & \checkmark &  &  & 74.69 & 57.58 \\
     & \checkmark &  & \checkmark &  & 75.56 & 59.34 \\
    
    \hline
     \checkmark & \checkmark & \checkmark &  &  & 80.96 & 66.54 \\
     \checkmark & \checkmark &  & \checkmark &  & 81.22 & 67.91 \\
      & \checkmark & \checkmark & \checkmark &  & 74.23 & 57.74 \\

    \hline
     \checkmark & \checkmark & \checkmark & \checkmark &  & 80.83 & 67.21  \\
     \checkmark & \checkmark & \checkmark &  & RN18 & 80.39 & 65.82 \\
     \checkmark & \checkmark & \checkmark &  & RN50 & 81.93 & 67.71\\
     \checkmark & \checkmark & \checkmark &  & RN101 & 81.80 & 67.79\\

    \hline
\end{tabular}
\label{Performance of models trained on different subsets of features}
\end{table}

Figure~\ref{final graph} shows the effect of human-interpretable privacy features in various combinations. We trained multiple MLPs with three hidden layers on features extracted by various deep learning models starting from ResNet18 and ending with more modern architectures such as Swin~\cite{b9} and ConvNeXt~\cite{b10}. To assess the usefulness of the proposed privacy-specific features, we trained similar MLPs on the concatenation of the eight privacy-specific features and features extracted from deep models. Models that consider eight privacy-specific features in addition to deep features consistently outperform models that use only deep-learning features by approximately 1 percentage point (p.p.). in F1 score and balanced accuracy.  We also applied an MLP with fewer hidden layers on the eight privacy-specific features only and achieved a performance of only 1.72p.p. worse than the best performance on deep features. Analysing the models' architectures, we see that base-size recent models with better predictions on ImageNet also provide better features for privacy classification than simpler ResNet101, even though they produce fewer features at the pre-last layer.

Features extracted by pre-trained models are supposed to be related to less abstract concepts as models were trained on concrete, non-subjective visual classification. We used pre-trained neural networks instead of fine-tuning them on the privacy classification to look at the usefulness of the privacy-specific human-interpretable features. However, fine-tuned models show much better results and can learn privacy-related features. For example, fine-tuned on privacy classification task ResNet101 with trained head (MLP with 3 hidden layers) outperforms previous models trained on deep and privacy-specific features significantly and does not benefit from human-interpretable features.
\begin{figure}[t]
    \centering
    \includegraphics[width=0.48\textwidth]{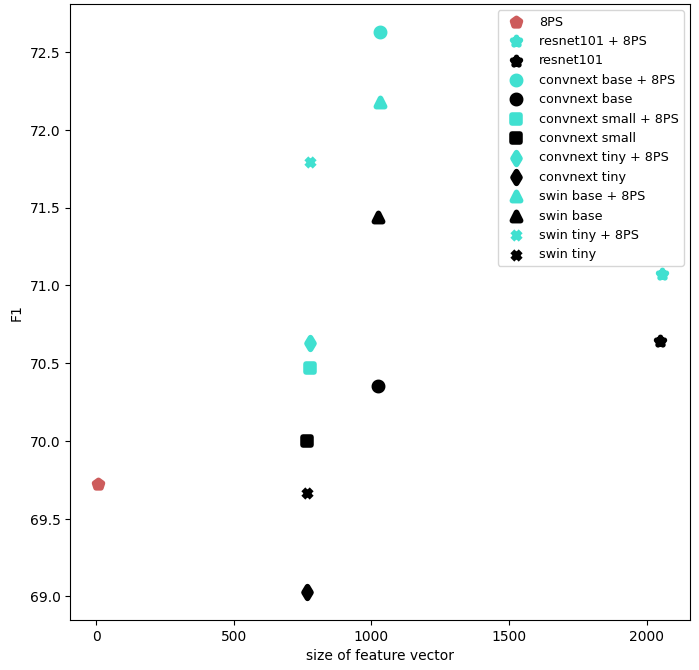}
    \caption{F1 score of MLP trained on features extracted by deep neural networks and/or privacy-specific human-interpretable features. Key -- 8PS: eight privacy-specific features } 
    \vspace*{-4mm}
    \label{final graph}
\end{figure}

Table~\ref{tab1} compares CNN, fine-tuned on privacy classification, our best MLP on the concatenation of deep and eight privacy-specific features and a model from Content-based Graph Neural network~\cite{b3}. The model, consisting of ResNet101 fine-tuned on the privacy classification dataset with an MLP head, provides the best performance. 
 MLP on the concatenation of eight privacy-specific features and features extracted from base-size ConvNeXt (ImageNet trained), achieves the best results among non-fine-tuned models. As expected, the performance of such a simple method is worse than the performance of current state-of-the-art methods. Predictions based on the feature-extracting model exhibit a 1.51p.p. decrease in F1 score in comparison to the fine-tuned ResNet, but outperform the graph-based model by 3.94p.p. in accuracy. The analysis of the datasets and 1p.p. improvement of the performance of models based on deep features with only eight features suggest that better privacy classification can be achieved by mimicking the human annotation process instead of developing more complex deep learning models, although more recent neural networks with better performance on ImageNet provide better features for privacy classification.
 \AtBeginEnvironment{tabular}{\normalsize}
\begin{table}[t]
\centering
\caption{Comparison of different models on PrivacyAlert. Key -- UBA: Unweighted Binary Accuracy, F1: F1 score, RN101-PA: ResNet101 fine-tuned on PrivacyAlert, CN-IN-8PS: MLP on ConvNeXt base (ImageNet trained) features +  8 privacy-specific features.}
\begin{tabular}{l c c }
\textbf{Models} & \textbf{\textit{UBA}}& \textbf{\textit{F1}} \\
\hline
RN101-PA & 87.10 & 74.14\\
CN-IN-8PS & 87.22 & 72.63 \\
GPA \cite{b3} & 83.28 & 59.80 \\
\hline
\end{tabular}
\label{tab1}
\end{table}
\section{Conclusion}
We analysed the human decision-making process of image privacy annotation on the PicAlert user-study dataset and the PrivacyAlert dataset to define a set of privacy-specific features that are human interpretable. These features increase the performance of deep models and, on their own, improve the image representation for privacy classification. We also showed that CNN models fine-tuned on the privacy classification task outperform non-fine-tuned feature-extracting methods. 
Future research includes defining an appropriate graph structure to account for the relationship of privacy-specific features in the prediction process.\\
\noindent {\bf Acknowledgement}. This work was supported by the CHIST-ERA programme through the Project GraphNEx, under UK EPSRC grant EP/V062107/1.

\bibliographystyle{ieeetr}
\bibliography{conference_updated}

\begin{thebibliography}{10}

\bibitem{b1}
A.~Tonge and C.~Caragea, ``Image privacy prediction using deep neural networks,'' {\em ACM Trans. Web}, vol.~14, apr 2020.

\bibitem{b2}
D.~Stoidis and A.~Cavallaro, ``Content-based graph privacy advisor,'' in {\em 2022 IEEE Eighth International Conference on Multimedia Big Data (BigMM)}, pp.~65--72, 2022.

\bibitem{b3}
G.~Yang, J.~Cao, Z.~Chen, J.~Guo, and J.~Li, ``Graph-based neural networks for explainable image privacy inference,'' {\em Pattern Recognition}, vol.~105, p.~107360, 2020.

\bibitem{b4}
S.~Zerr, S.~Siersdorfer, J.~Hare, and E.~Demidova, ``Privacy-aware image classification and search,'' in {\em Proceedings of the 35th International ACM SIGIR Conference on Research and Development in Information Retrieval}, SIGIR '12, (New York, NY, USA), p.~35–44, Association for Computing Machinery, 2012.

\bibitem{b5}
G.~Ayci, P.~Yolum, A.~{\"O}zg{\"u}r, and M.~{\c{S}}ensoy, ``Explain to me: Towards understanding privacy decisions,'' {\em arXiv preprint arXiv:2301.02079}, 2023.

\bibitem{b6}
C.~Zhao, J.~Mangat, S.~Koujalgi, A.~Squicciarini, and C.~Caragea, ``Privacyalert: A dataset for image privacy prediction,'' {\em Proceedings of the International AAAI Conference on Web and Social Media}, vol.~16, p.~1352–1361, May 2022.

\bibitem{b7}
S.~Zerr, S.~Siersdorfer, and J.~Hare, ``Picalert! a system for privacy-aware image classification and retrieval,'' in {\em Proceedings of the 21st ACM international conference on Information and knowledge management}, pp.~2710--2712, 2012.

\bibitem{b9}
Z.~Liu, Y.~Lin, Y.~Cao, H.~Hu, Y.~Wei, Z.~Zhang, S.~Lin, and B.~Guo, ``Swin transformer: Hierarchical vision transformer using shifted windows,'' in {\em Proceedings of the IEEE/CVF international conference on computer vision}, pp.~10012--10022, 2021.

\bibitem{b10}
Z.~Liu, H.~Mao, C.-Y. Wu, C.~Feichtenhofer, T.~Darrell, and S.~Xie, ``A convnet for the 2020s,'' in {\em Proceedings of the IEEE/CVF conference on computer vision and pattern recognition}, pp.~11976--11986, 2022.

\bibitem{b8}
B.~Zhou, A.~Khosla, A.~Lapedriza, A.~Torralba, and A.~Oliva, ``Places: An image database for deep scene understanding,'' {\em arXiv preprint arXiv:1610.02055}, 2016.

\bibitem{b11}
G.~Jocher, A.~Chaurasia, A.~Stoken, J.~Borovec, {\em et~al.}, ``{ultralytics/yolov5: v7.0 - YOLOv5 SOTA Realtime Instance Segmentation},'' Nov. 2022.

\bibitem{b12}
Google, ``Safesearch \uppercase{API}. \uppercase{A}vailable online: https://cloud.google.com/vision/docs/detecting-safe-search.''
\newblock accessed on 24 December 2022.

\end{thebibliography}

\vspace{12pt}
\color{red}
\end{twocolumn}
\end{document}